# Keeping Teams in the Game: Predicting Dropouts in Online Problem-Based Learning Competition

**Aditya PANWAR[a*], Ashwin TS[a], Ramkumar RAJENDRAN[a] & Kavi ARYA[a]**
[a]*IDP in Educational Technology, Indian Institute of Technology Bombay, INDIA*
adityapanwar@iitb.ac.in

**Abstract:** Online learning and MOOCs have become increasingly popular in recent years, and the trend will continue, given the technology boom. There is a dire need to observe learners' behavior in these online courses, similar to what instructors do in a face-to-face classroom. Learners' strategies and activities become crucial to understanding their behavior. One major challenge in online courses is predicting and preventing dropout behavior. While several studies have tried to perform such analysis, there is still a shortage of studies that employ different data streams to understand and predict the drop rates. Moreover, studies rarely use a fully online team-based collaborative environment as their context. Thus, the current study employs an online longitudinal problem-based learning (PBL) collaborative robotics competition as the testbed. Through methodological triangulation, the study aims to predict dropout behavior via the contributions of Discourse discussion forum 'activities' of participating teams, along with a self-reported Online Learning Strategies Questionnaire (OSLQ). The study also uses Qualitative interviews to enhance the ground truth and results. The OSLQ data is collected from more than 4000 participants. Furthermore, the study seeks to establish the reliability of OSLQ to advance research within online environments. Various Machine Learning algorithms are applied to analyze the data. The findings demonstrate the reliability of OSLQ with our substantial sample size and reveal promising results for predicting the dropout rate in online competition. Overall, the study contributes to online learning by addressing the need to understand and predict dropout behavior in online courses. The study's methodological triangulation, involving qualitative interviews, provides insights into such contexts' unique dynamics and challenges by utilizing a fully online team-based collaborative environment.

## 1. Introduction & Related Literature

There has been a tremendous increase in online courses worldwide for Undergraduate students in engineering and all other fields. COVID-19 has further triggered MOOC enrolment numbers to an exponential level (Baudo & Mezzera, 2021). However, the most prominent drawback of online courses is their high attrition rate; the completion rate usually falls between 2-10% (Maxwell et al., 2018). The success and learning in online courses may be directly attributed to student performance, which can be further attributed to several key factors like content interaction, student satisfaction (Zimmerman, 2012), engagement & activity with course material (Soffer & Cohen, 2019), and motivation (Loizzo, Ertmer, Watson, & Watson, 2017), to list a few. Many such factors can be observed or assessed in a traditional classroom with limited participants. The instructor can guide, adapt and help the 'failing' students towards success. All these factors become even more critical when dealing with the online setting where the significant learner interactions are via the resources/content and maybe a discussion or query forum. This necessitates exploring the reasons for dropouts to alleviate this issue and provide an opportunity to inform the instructors to adapt the curriculum/task design, motivate, provide feedback, and devise other ways to keep success high. It is known that there are numerous benefits of online courses, such as rich content, online asynchronous learning, and low cost. Still, despite that, they continue to suffer from significant dropout rates, hampering pedagogical and economic goals (Quadri & Shukor, 2021). Predicting students' likelihood to complete (or not to complete) a MOOC course, especially from the early weeks, has been one of the hottest research topics in the area of learning analytics (Alamri, Alshehri, Cristea, Pereira, Oliveira, Shi, & Stewart, 2019). Predictive models provide timely information about learners at risk of dropout to inform interventions. Instructors and learners can benefit from the results of the predictive models (Moreno-Marcos, Alario-Hoyos, Muñoz-Merino, & Kloos, 2018), which can help modify and improve course content and pedagogy to reduce dropouts (Moreno-Marcos, Muñoz-Merino, Maldonado-Mahauad, Pérez-Sanagustín, Alario-Hoyos, & Delgado Kloos, 2020). There is a huge potential for data analytics on students' learning processes and outcomes in higher education (Aldowah, Al-Samarraie, & Fauzy, 2019). Learning analytics is the measurement, collection, analysis, and reporting of data about learners and their contexts to understand and optimize learning and the environments in which it occurs (Siemens & Long, 2011). Students' interaction with online content is crucial and utilized in many research studies. Students' active participation in forums was found to enhance overall performance. The completion level was also closely linked to prior online experience and educational attainment (Meneses & Marlon, 2020). Students' interactions are an important source for understanding the students' behavioral patterns in an online learning process. Other researchers have also claimed that students' social and personal information, such as gender, age, or place, significantly impacted their performance and learning outcomes in general (Mubarak, Cao, & Zhang, 2020). Success in an online course requires a high level of discipline and self-direction, and proper time management to complete the assignments. Thus, certain learning strategies must be developed, and the instructors should encourage the development. Peer Learning should also be encouraged by forming project/study groups wherever possible. Though many prediction models have been developed, they usually don't consider high-level factors like Self-Regulated Learning (SRL), which can greatly impact the learners' success. Prior works have also shown that the lack of SRL skills can be an important factor that leads to failure and dropout in MOOCs (Terras & Ramsay, J. 2015). In a study conducted by a research group at Harvard (Whitehill et al., 2015), multinomial logistic regression was employed to identify students at risk of dropping out of a MOOC. Other algorithms, including Logistic and Linear Regression, Forest Regression, Decision Tree, etc., have also been employed (Liao et al., 2019). It has been noticed that although a decent amount of work has been done in predicting dropouts, they mostly focus on a single stream of data, like learners' interaction with content. The dropout phenomena are studied quantitatively, whereas research emphasizes that dropouts can be highly qualitative and complex (Simpson, 2010). Dropout studies thus lack information on important socio-psychological causes and contingencies (e.g., academic workload, personal experiences, and other commitments). SRL is found to be one of the major factors in online learning. Yet, it is not fully utilized for prediction, so a need is there to use multiple data sources [learner interaction in an online forum, SRL Strategies, Qualitative Studies, learner profile, etc.] as combinations and predictors. There has been quite a lot of research going on in the field of motivation and Learning Strategies, yet there is a research gap in purely online learning scenarios. SRL is usually studied through questionnaires; the Motivated Strategies for Learning Questionnaire (MSLQ) has been widely used. However, it was developed in 1991 and is primarily suited to face-to-face classrooms (Zhou & Wang, 2021). Efforts have been made to modify MSLQ part B for Distance Education (Meijs, Neroni, Gijselaers, Leontjevas, Kirschner, & de Groot, 2019). Still, not many descriptive studies have been conducted to establish the validity. Several other instruments like MSLQ (basic), MSLQ modified, and Self-regulated Online Learning Questionnaire SOL-Q (Jansen, van Leeuwen, Janssen, Kester, & Kalz, 2017), Technology Innovation Questionnaire, NSSE (Pascarella, Seifert, & Blaich, 2010), Educause Student Technology Survey (Dahlstrom, & Bichsel, 2014), and Self-Regulation for Learning Online (SRL-O) (Broadbent, Panadero, Lodge, & Fuller-Tyszkiewicz, 2022) are also sporadically used in studies. Online Self-Regulated Learning Questionnaire or OSLQ (Barnard, Lan, To, Paton, & Lai, 2009; Bruso & Stefaniak,

2016) is one such instrument developed solely for the online environment and accommodates the most important scales from the literature like time management, task strategies, etc. However, many more studies with large sample sizes (Rufini, Fernandes, Bianchini, & Alliprandini, 2021) using OSLQ are warranted.

Several dropout prediction models are created, but they are carried out as posthoc analysis, which fails to 'anticipate' the dropouts; early predictions and sequentially suitable intervention strategies are thus the need of the hour. Moreover, the research almost always involves individual participants, so there is a huge shortage of team or collaborative research. The ability to predict a team's performance can have a strong pedagogical potential that has not yet been explored sufficiently in the related literature (Giannakas, Troussas, Voyiatzis, & Sgouropoulou, 2021). Furthermore, when dealing with questionnaires, issues related to self-reported data are always present; this can be somewhat circumvented by conducting longitudinal studies and qualitative interviews. Through this study, we are thus trying to tackle these gaps via

1. Collecting self-reported self-regulated learning (SRL) data through a questionnaire and learner activity data from the discussion forum,
2. Analyzing the SRL data to understand its implications within a team-based problem-based learning (PBL) environment,
3. Developing Early Prediction models specifically designed for an online team-based setting to reduce attrition,
4. Augmenting the investigation with qualitative interviews conducted with teams/team members.

Based on these aims, we propose the following Research Questions:

**RQ1**: How reliably can the Online Strategies for Learning Questionnaire (OSLQ) demonstrate the level of Self-Regulated Learning (SRL) in the context of a fully online problem-based learning (PBL) competition?

**RQ2**: How effective are the early prediction models based on discussion forum interactions and the OSLQ in anticipating and preventing team dropouts in the online PBL environment?

In the upcoming sections of this study, we will describe the study environment and discuss the data collection methods, specifically the utilization of the OSLQ (SRL Questionnaire) and the incorporation of Discourse discussion forum activity data like *Topics visited, Posts read, likes received, and given*. We will then proceed to build our prediction models and present the results. Finally, we will conclude with a comprehensive discussion section that analyzes and interprets the findings, contextualizing them within the broader scope of our research and exploring their implications.

## 2. Study Environment

To address our RQs, we use e-Yantra Robotics Competition (**eYRC**) as our study environment. One of the authors worked within e-Yantra, which allowed the use of eYRC. e-Yantra is an initiative hosted in IIT Bombay (in CSE Dept.) with the support of the Ministry of Education (MoE) to spread a wide variety of technical skills across the college. e-Yantra's initiatives complement the existing education system with a "problem-solving" culture through project-based learning (PBL) approach using Simulators & Robotic Kits. eYRC focuses on building not only "HARD" [STEM, Image processing, control systems, ML, embedded systems, Functional programming, etc.] but "SOFT" [teamwork, time-management, dealing with failure, etc.] skills too. eYRC is unique because it is really a MOOC-style competition that trains participants in complex engineering while competing. The students participate as a team of two to four student members from any year and any branch. Participating teams try to solve gamified real-world problem statements called **Themes** and are abstracted on a static arena. Incepted in 2012 (eYRC-2012), the competition occurs online in two stages, as shown in Figure 1. Stage 1 is usually simulator/programming based, comprising three tasks-Task 0, Task 1, and Task 2, occurring during September-November. Teams are then selected based on their performance and advance into Stage 2, which involves hardware and learnings from Stage 1. It consists of Task 3, Task 4, Task 5, and Task 6 conducted between December-March. Every team member has access to all resources, the discussion forum, eYRC website (*that hosts questionnaires, feedback forms, etc., needed to be filled out individually by each member*). The submission of each task is, however, team-based rather than individual. Each theme comprises tasks and constant feedback provided through various online discussion forums, including Discourse, Discord, and Google Hangouts. The selection of eYRC as the research setting is influenced by its unique attributes as a comprehensive online collaborative robotics competition. Furthermore, segmenting problem statements into distinct deadline-driven tasks within eYRC offers an opportune platform for applying algorithms in early prediction endeavors. First, we gather individual members' SRL questionnaire responses via the website and each member's interaction data using the Discourse discussion forum. It is crucial to consider that interaction data should be analyzed within the specific time frame assigned to each task, as each task has its own deadline. Semi-structured interviews are also conducted with the eYRC participants for our study to delve deeper into dropout behavior. Then, using methodological triangulation, we try to achieve the study goals.

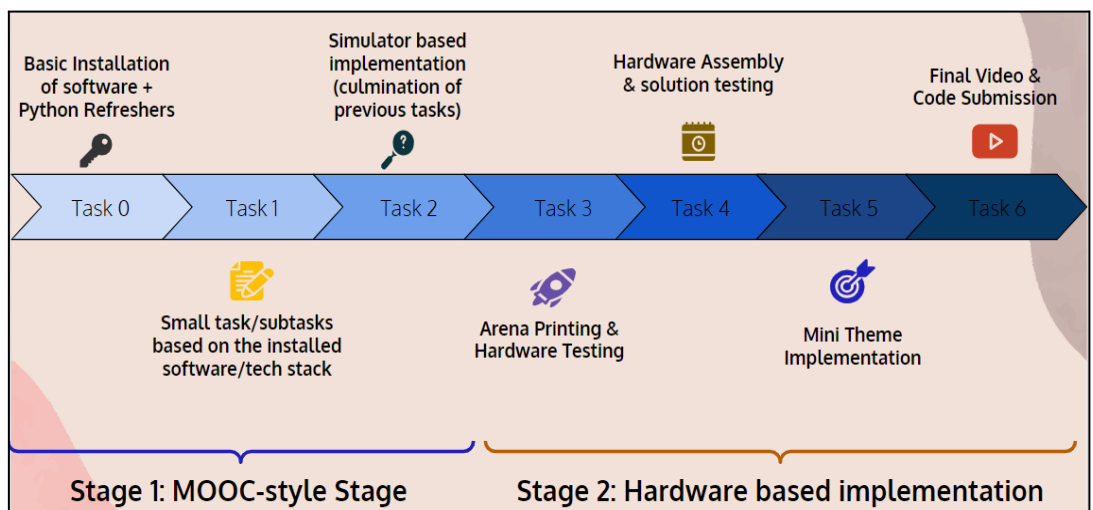

*Figure 1*. eYRC Structure

## 3. Research Design & Methodology

Our study employs methodological triangulation that uses more than one kind of method to study a phenomenon. It is beneficial in confirming findings, more comprehensive data, increasing validity, and enhancing understanding of studied phenomena (Bekhet & Zauszniewski, 2012). This framework underpins our use of the SRL questionnaire, Discussion forum, and qualitative interviews, as detailed in the following sections.

### 3.1 *Learning Strategies Questionnaire*

For this study, OSLQ: Online Strategies for Learning Questionnaire, which has six sub-scales, namely, **Goal setting, Environment Setting, Task Strategies, Time Management, Help Seeking, and Self Evaluation,** is used. OSLQ is rated via a 5-point Likert Scale, with 1 being *Strongly Disagree* and 5 being *Strongly Agree*. OSLQ is a developed model; hence no Exploratory Factor Analysis (EFA) will be necessary; Confirmatory Factor Analysis (CFA) will be required to establish reliability and address RQ1. CFA enables us to establish construct validity by testing the relationship between observed factors (question items) and latent factors (subscales) proposed by the model. This step is necessary to ensure that subscale scores within the SRL framework can be distinguished and thus reported separately. Jamovi being open-source and cloud-based, was tested for performing the CFA. OSLQ was administered to participants via the web portal with barely any modifications to **10011 individual participants representing 2610 teams** registered entering Stage 1 of the competition to attempt Task 0 [*basic software installation and a supplementary task*]. **4907 individual participants, representing 1520 teams,** had filled out the questionnaire, and further analysis was performed on this dataset.

### 3.2  Discourse Discussion Forum Data

To address RQ2, Discourse (an open-source discussion forum) was utilized in the competition. Participants were enrolled here to access various resources, announcements and post their queries. Discourse readily provides forming "groups," where an instructor can add only certain participants to a certain group relevant to their 'theme.' Six separate groups were formed for six Themes in the analyzed year (2022-23). The total number of individual participants from these six themes was 7604 representing 2214 teams.

Discussion forum learner activity was gathered using the Data Explorer feature of Discourse, where full-fledged Structured Query Language (SQL) queries can be written. Queries like Poll Statistics, Lurkers, High Likers, and many more can be explored by simply running them on the cloud. We also leveraged SQL queries to extrapolate each group's data using their unique *group_id*. This data was first separately collected for each theme group based on the tasks' timeline and then later combined for overall analysis. Major features identified from the data were ***Topics Entered, Posts Count, Likes Given, and Likes Received*** (see Appendix A)[1]*.* Since these values were individual-based, they had to be further processed to fit the team-based style of the competition. The **MAX** values of each feature were finalized because of the nature of the competition that even if one team member "visits" the resources, it can be constituted to the whole team. Hence, we finally utilized four features for further analysis and model building for each team: ***Max Likes received, Max Likes Given, Max Topics Entered, and Max Posts Read.*** Furthermore, the time interval (deadline) for the above features between tasks was also crucial since we aim for early predictions. For example, for task1 dropout prediction, we only used the features extracted before the task1 deadline, i.e., Task 0 and Task 1. A similar process was followed for other predictions.

#### 3.2.1  Building Machine Learning Models

All algorithms were run in Python using Google Colab. Since the OSLQ data is in Likert Scale, it was first standardized using suitable Python libraries.

#### 3.2.1.1.  Procedure

To build our model, we first performed data pre-processing. An overview of the steps:
1. OSLQ data with aggregate (1520 teams) added with the team's Discourse activity.
2. After this, each team's task submission data was added. Task submission data either scored (value >=0) or value NULL, representing a team not submitting task
3. NULL value and scores were suitably standardized to 0 (Not Submitted) and 1 (Submitted) using label encoders. Performance or scoring is not considered to keep the model simple (see Appendix B[1] for the Merged dataset description).
4. As our environment resembles a MOOC model, there was a lot of imbalance in the dataset. An example of this is shown in Figure 2; teams selected in Task 0 of the competition.
5. We used SMOTE as the technique to balance the data. SMOTE stands for Synthetic Minority Over-sampling Technique, and it works by generating synthetic examples of the minority class by interpolating between existing examples in the minority class. Similar data balancing is also used to balance the task-wise data since we focus on early prediction. Figure 3 shows Task 0 composition after SMOTE.

---
[1] [Appendix Document](#)

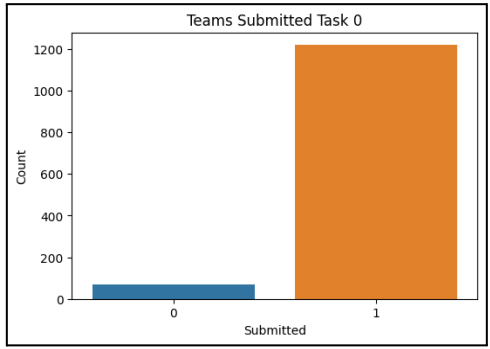

*Figure 2.* Task 0 Imbalanced Dataset

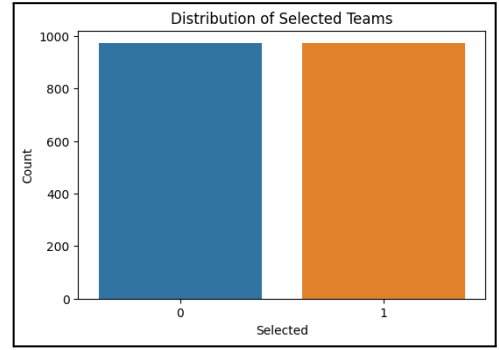

*Figure 3.* Task 0 Balanced Dataset

6. We then employed competing ML methods: Logistic Regression, Decision Trees, and Random Forest (see Appendix B[2] for details).
7. After building the above models based on a training and test set division of 80%/20%, we also evaluated the models for all performance metrics: accuracy, precision, recall, F1-score, and AUC-ROC (see Appendix B[2] for details).

In conclusion, the data pre-processing for our machine learning models involved combining OSLQ data with team Discourse activity and incorporating task submission data. We standardized the task submission data to represent submission status. Due to our MOOC-like environment, we encountered dataset imbalance, which we addressed using the SMOTE and employed Logistic Regression, Decision Trees, and Random Forest as competing ML methods. Finally, their performance was evaluated using various metrics such as accuracy, precision, recall, F1-score, and AUC-ROC.

### 3.3 *Qualitative Data*

To establish further reliability, a qualitative study was conducted with eYRC participants that were selected from a pool of eYRC 2021-22 (last year's version) who were not able to perform well in Stage 1 (dropped out before Task 2) but participated again in the current version (eYRC 2022-23) and cracked the Stage 1 to enter the hardware-based Stage 2. **Eight participants** representing **six teams** were invited to participate and consented to a study. The overarching objective was to comprehend the team's view of the discussion forum, unravel the causes of early dropout, analyze the role of discussion forum activities, elucidate their contributions, and triangulate this with our Early Prediction models. These interviews were conducted online over Google Meet. Audio and video were both recorded for analysis purposes. To facilitate the discussion with the participants, guiding questions were provided. Each group member was asked these initial questions to further the conversation (see Appendix C)[2]. Otter (a speech-to-text app) was employed, and handwritten notes were taken as a backup.

### 4. Results and Analysis

In this section, we present the results and analysis of the OSLQ (Self-Regulated Learning Questionnaire) data addressing RQ1. Descriptive statistics and Cronbach's Alpha were used to assess the internal consistency of the questionnaire. We also present the CFA results and the adequacy of the model fit to the data as a satisfactory model fit is crucial for the reliability of the questionnaire. We then present the performance and evaluation of our Early Prediction models to address RQ2. Finally, we discuss the analysis of interviews.

---

[2] Appendix Document

## 4.1 OSLQ Data Analysis

To answer our RQ1, initially, Descriptive Statistics were performed on the resulting data from OSLQ **(n = 4907)**. Internal consistency for all six subscales was determined using *Cronbach's Alpha*, as shown in Table 1. All subscales lie in an acceptable range of (0.70 to 0.95) of Cronbach's Alpha and are sufficiently reliable for analysis.

Table 1. *OSLQ Internal Consistency*

| Sub Scales | Cronbach's Alpha (α) |
|---|---|
| Goal Setting | 0.864 |
| Environment Setting | 0.814 |
| Task Strategies | 0.729 |
| Time Management | 0.812 |
| Help Seeking | 0.799 |
| Self Evaluation | 0.846 |

In performing the CFA, five statistics reflecting fit were reported: the chi-square goodness of fit statistic ($\chi 2$); the ratio of chi-square statistic to degrees of freedom ($\chi 2/df$); the Comparative Fit Index (CFI); the Tucker Lewis Index (TLI) [also known as the Non-Normed Fit Index (NNFI)]; and the root mean square error of approximation (RMSEA). The model fit results are shown in Table 2 and Table 3 below.

Table 2. *Exact Model Fit*

| $\chi^2$ | df | p |
|---|---|---|
| 2924 | 237 | < .001 |

Table 3. *Fit Measures*

| | | | RMSEA 90% CI | |
|---|---|---|---|---|
| CFI | TLQ | RMSEA | Lower | Upper |
| 0.948 | .94 | .0531 | .0514 | 0.548 |

Cronbach's Alpha confirmed the reliability of the questionnaire's subscales. CFA established the validity of the questionnaire, ensuring that subscale scores within the SRL framework could be reliably distinguished. The fit statistics of the CFA were examined, and a good fit was found. Descriptive statistics, CFA loadings, and Model Fit Indices description are added in Appendix D[3].

## 4.2 Machine Learning Models Performance

From the initial pool of 1520 teams, a dataset comprising 1290 teams was selected for addressing RQ2. Initial model performance was evaluated by constructing models without oversampling using Logistic Regression, Decision Trees, and Random Forest. Subsequently, the model was refined by incorporating minority oversampling for task0, task1, task2, and stage2 as target variables—early Prediction utilized Discussion Forum activities data before each task's deadline. The results are presented in Table 4, Table 5, and Table 6.

Table 4. *Logistic Regression report for task0, task1, task2 and stage2 prediction*

| | Precision | Recall | F1 Score | AUC-ROC | Accuracy |
|---|---|---|---|---|---|
| task0 | 0 | 0 | 0 | 0.49 | 0.95 |
| | 0.96 | 1 | 0.98 | | |

---

[3] Appendix Document

|  | | | | | |
|---|---|---|---|---|---|
| task1 | 0.61 | 0.43 | 0.5 | 0.63 | 0.7 |
|  | 0.96 | 1 | 0.98 | | |
| task2 | 0.85 | 0.94 | 0.9 | 0.83 | 0.86 |
|  | 0.89 | 0.73 | 0.8 | | |
| stage2 | 0.92 | 0.93 | 0.93 | 0.8 | 0.88 |
|  | 0.71 | 0.69 | 0.7 | | |

Table 5. *Decision Tree model report for task0, task1, task2 and stage2 prediction*

|  | Precision | Recall | F1 Score | AUC-ROC | Accuracy |
|---|---|---|---|---|---|
| task0 | 0.11 | 0.45 | 0.18 | 0.64 | 0.82 |
|  | 0.97 | 0.94 | 0.95 | | |
| task1 | 0.5 | 0.58 | 0.54 | 0.64 | 0.65 |
|  | 0.75 | 0.69 | 0.72 | | |
| task2 | 0.83 | 0.84 | 0.83 | 0.77 | 0.79 |
|  | 0.69 | 0.72 | 0.7 | | |
| stage2 | 0.94 | 0.91 | 0.93 | 0.84 | 0.88 |
|  | 0.68 | 0.78 | 0.73 | | |

Table 6. *Random Forest model report for task0, task1, task2 and stage2 prediction*

|  | Precision | Recall | F1 Score | AUC-ROC | Accuracy |
|---|---|---|---|---|---|
| task0 | 0.27 | 0.36 | 0.31 | 0.65 | 0.93 |
|  | 0.96 | 0.99 | 0.97 | | |
| task1 | 0.58 | 0.65 | 0.61 | 0.70 | 0.71 |
|  | 0.78 | 0.83 | 0.80 | | |
| task2 | 0.89 | 0.87 | 0.88 | 0.84 | 0.85 |
|  | 0.79 | 0.81 | 0.80 | | |
| stage2 | 0.97 | 0.91 | 0.94 | 0.90 | 0.91 |
|  | 0.71 | 0.88 | 0.79 | | |

The study employed machine learning models - Logistic Regression, Decision Tree, and Random Forest - to predict task completion and stage advancement in the e-Yantra Robotics Competition. Results indicated varied model performance across tasks and stages. Notably, the Random Forest model consistently outperformed others, demonstrating higher accuracy and AUC-ROC values. For instance, in predicting Task 0 (task0), Random Forest achieved an accuracy of 0.93 and AUC-ROC of 0.65. Similarly, Task 1 (task1) prediction achieved an accuracy of 0.71 and an AUC-ROC of 0.70. In Task 2 (task2) and Stage 2 (stage2) predictions, Random Forest maintained competitive accuracy and AUC-ROC values, solidifying its predictive prowess.

### 4.3 Qualitative Data Analysis

Through the analysis of the interviews, several themes and issues emerged. The participants in the study found the Discourse Discussion Forum highly effective, being more active on it this year, assessing queries before posting. Participants found it beneficial to observe the progress of other participants in the forum. They felt more satisfied and reassured when they discovered that others were facing similar challenges or getting stuck at similar points in the competition while acknowledging helpful solutions with likes and appreciating the forum's structure and navigation. Additionally, they demonstrated improved time management between academics and competition, consistent effort in completing tasks, and reported the benefits of prior participation. Participants also identified issues about the importance of diligently choosing team members and being shy to post their queries in their previous participation, owing to how they'll be perceived on the forum.

## 5. Conclusion

This study aimed to predict dropout behavior in an online collaborative PBL competition using a methodological triangulation approach. The study combined data from participating teams' Discourse discussion forum activities, a self-reported Online Learning Strategies Questionnaire (OSLQ), and qualitative interviews to enhance the results. OSLQ descriptive results revealed that subscales Goal Setting (µ=4.06), Environment Setting (µ=4.1525), Help Seeking (µ=4.03), and Self Evaluation (µ=4.06) were highly evidenced, while Task Strategies (µ=3.8225) and Time Management (µ=3.85) were on the lower side. A reliability analysis of each subscale was performed, and all subscales were found to be within the acceptable range of 0.70 and above, making OSLQ valid and consistent. Using CFA, we established the reliability of OSLQ using a large sample size from various institutions with participants representing various disciplines of Undergraduate Studies in a team-based dynamics. The qualitative interviews also pointed toward participants' perception of the importance of Time Management as a success indicator. Instructors can thus create some interventions right at the beginning of the competition to encourage participants. For building machine learning early prediction models, we kept the general aim of keeping the models fairly simple and used discussion forum features. The models were built till Stage 1 (or Stage 2 selection) as we focus on Early Predictions. The performance was poor at the start of the competition (task0) but gradually improved. Random Forest and Decision Trees performed better than the Logistic Regression models. Decision Trees and Random Forest seemingly worked well as they incorporated non-linear relationships between the variables. These findings highlight the potential of machine learning models, particularly Random Forest, in predicting task completion and stage advancement in the online Competition. The results contribute to our understanding of the factors influencing the team's potential early dropouts and can inform the design of interventions to enhance their learning outcomes. The findings of this study have implications for online course design, instructor interventions, and learner support strategies. By identifying teams at risk of dropout early on, instructors can offer timely support, adaptive feedback, and personalized interventions to improve success. However, it is essential to acknowledge some limitations of this study. The self-reported nature of the OSLQ data introduces potential biases and may not capture the full complexity of learners' strategies and behaviors. In the ongoing work, we are exploring the role of individual participants' data like Year of Study, Prior participation, Engineering Branch, Gender, and other demographic variables that could provide even more insights on team-based research in the online scenario. In conclusion, the contribution of this ongoing study is multifaceted; first, it cements the OSLQ's validity and reliability through a large sample size which has been missing in the literature. Secondly, it contributes to the growing body of research on predicting dropout behavior in online courses, especially in a fully online longitudinal environment that is also collaborative.